\pdfoutput=1

\documentclass[10pt,twocolumn,letterpaper]{article}

\usepackage[accsupp]{axessibility}

\usepackage{iccv}
\usepackage{times}
\usepackage{epsfig}
\usepackage{graphicx}
\usepackage{amsmath}
\usepackage{amssymb}
\usepackage{enumitem}
\usepackage{breqn}
\usepackage{mathtools}
\usepackage{authblk}

\newcommand{\OURNAME}{ZFlow\xspace}

\newcommand{\GAF}{Gated Appearance Flow\xspace}
\newcommand{\DGP}{Dense Geometric Priors\xspace}
\newcommand{\DGAR}{Dense Garment-Agnostic Representation\xspace}
\newcommand{\IUV}{IUV Priors\xspace}

\newcommand{\GAFshort}{GAF\xspace}

\newcommand{\DGARshort}{DGAR\xspace}

\makeatletter
\newcommand{\printfnsymbol}[1]{%
  \textsuperscript{\@fnsymbol{#1}}%
}
\makeatother

\usepackage[pagebackref=true,breaklinks=true,letterpaper=true,colorlinks,bookmarks=false]{hyperref}

\iccvfinalcopy 

\def\iccvPaperID{10355} 

\ificcvfinal\fi

\begin{document}

\title{\OURNAME: Gated Appearance Flow-based Virtual Try-on with 3D Priors}

\author[2]{Ayush Chopra\thanks{equal contribution}\thanks{work done while working at Adobe MDSR Lab}}
\author[3]{Rishabh Jain \printfnsymbol{1}\thanks{work done as part of Adobe MDSR internship}}
\author[1]{Mayur Hemani}
\author[1]{Balaji Krishnamurthy}

\affil[1]{Media and Data Science Research Lab, Adobe}
\affil[2]{Media Lab, Massachusetts Institute of Technology}
\affil[3]{BITS Pilani}

\maketitle
\ificcvfinal\thispagestyle{empty}\fi

\begin{abstract}
Image-based virtual try-on involves synthesising perceptually convincing images of a model wearing a particular garment and has garnered significant research interest due to its immense practical applicability. Recent methods involve a two stage process: i) warping of the garment to align with the model ii) texture fusion of the warped garment and target model to generate the try-on output. Issues arise due to the non-rigid nature of garments and the lack of geometric information about the model or the garment. It often results in improper rendering of granular details. We propose \OURNAME, an end-to-end framework, which seeks to alleviate these concerns regarding geometric and textural integrity (such as pose, depth-ordering, skin and neckline reproduction) through a combination of gated aggregation of hierarchical flow estimates termed \textit{\GAF}, and dense structural priors at various stage of the network. \OURNAME achieves state-of-the-art results as observed qualitatively, and on quantitative benchmarks of image quality (PSNR, SSIM, and FID). The paper presents extensive comparisons with other existing solutions including a detailed user study and ablation studies to gauge the effect of each of our contributions on multiple datasets.
\end{abstract}


\section{Introduction}
With recent socio-cultural events accelerating the shift towards online commerce, there is an increasing interest in providing smart and intuitive experiences~\cite{sievenet, tagging_icip, ayushcontext, cvpr_attr, chopra2019powering, Lang_2020_CVPR} that can compensate for the lack of in-store interaction. Virtual try-on is concerned with the visualization of clothes in a personalized setting and is of great importance to a plethora of real world applications. While attractive even before the renaissance of deep learning~\cite{tanaka2009texture, hauswiesner2013virtual, ehara2006texture}, recent advances in generative networks have inspired researchers to pursue image-based virtual try-on~\cite{acgpn, sievenet, cpvton, han2019clothflow, vtnfp}, based solely on RGB images, by formulating the problem as that of conditional image synthesis.

\begin{figure}
\begin{center}
  \includegraphics[width=\linewidth]{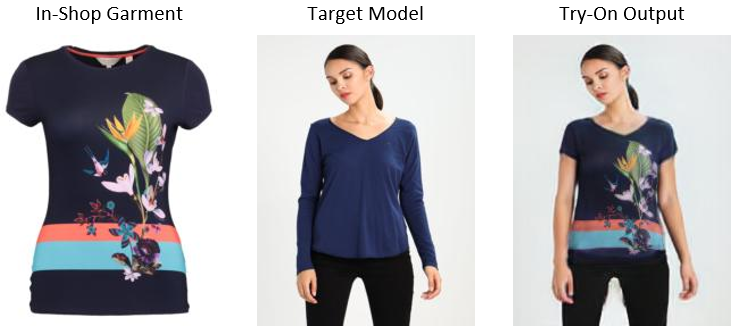}
\end{center}
  \caption{Image-based virtual try-on involves
synthesizing a \textit{try-on output} where the \textit{target model} is wearing the \textit{in-shop garment} while other characteristics of the model and garment are preserved. The above output is generated by our proposed method \OURNAME} 
\label{fig:tryon-problem}
\end{figure}

Given as input the images of an \textit{isolated in-shop garment} and a \textit{target model}, the objective of image-based virtual try-on is to synthesise a perceptually convincing new image (referred to as the \textit{try-on output}) where the target model is wearing the in-shop garment (Figure~\ref{fig:tryon-problem}). Recent methods employ a two step process consisting of: a) \textit{warping} of in-shop garment to align with pose and body shape of the target model and, b) \textit{texture fusion} of the warped garment and target model images to generate the try-on output. A successful try-on experience depends upon synthesizing a sharp, realistic image that preserves the textural and geometric integrity of both the garment and model. Issues arise from improper warping or incorrect texture fusion due to the non-rigid nature of garments and the lack of understanding of the 3D geometry of the garment and the model. This results in unconvincing rendering of granular clothing details. Alleviating these concerns is the focus of this work.

Recent research ~\cite{viton, cpvton, sievenet, acgpn} has been directed towards these challenges. ~\cite{viton, cpvton} proposed thin-plate spline (TPS) based warping of the garment image. ~\cite{sievenet, acgpn} improve the stability of TPS warping via multi-stage cascaded parameter estimation, and second order difference constraints respectively. However, TPS based warping leads to inaccurate transformation estimation when large geometric deformation is required, since each parameter defines the spatial deformation for a coarse block of pixels. To address this issue, ~\cite{han2019clothflow} proposes to use dense, per-pixel appearance flow ~\cite{appearanceFlow} prediction to spatially deform the garment image. But owing to the high degree of freedom and the absence of proper regularization, this method often causes drastic deformation during warping resulting in significant textural artefacts. To address both issues - the inability of TPS to handle large deformations, and over-warping with appearance flows - we introduce \textit{\GAF} (GAF) which regularizes per-pixel appearance flow by aggregating candidate flow estimates predicted across multiple scales. 

Next, for improving texture fusion, especially the issue of bleeding colors, \cite{sievenet, acgpn} propose to use an \textit{apriori} estimate of target clothing segmentation for the try-on output as conditioning. However, this method results in ambiguities in depth perception and body-part ordering because of the absence of 3D geometric priors. This is prominently visible in the generation of necklines, and handling cases with occlusion. For example, part of the garment that should go behind the neck appears in the front. To encode the 3D geometry information, we combine UV projection maps with dense body-part segmentation (obtained via DensePose~\cite{densepose}) as priors during warping and texture fusion. 

Our contributions can be summarized as follows:
\begin{itemize}[noitemsep]
    \item We propose ZFlow, an end-to-end try-on framework, that utilizes gated appearance flow estimates and dense geometric priors to render high quality try-on results.
    \item We present extensive quantitative and qualitative comparisons as well as a detailed user study to show significant improvement over state-of-the-art methods.
    \item We present ablation studies to analyse impact of different design choices in \OURNAME. We further reinforce the efficacy of \textit{GAF} by adapting it to improve state-of-the-art for human pose transfer.
\end{itemize}
\section{Related Work}
~\label{sec:related-work}
\textbf{Virtual Try-On} 
Progress in deep learning has motivated 2D image-based try-on as a scalable alternative to older methods (\cite{sekine2014virtual, pons2017clothcap, tanaka2009texture, zhou2012image}) that used 3D scanners for virtual fitting of clothing items. Most of these new 2D image-based methods \cite{viton, cpvton, acgpn, sievenet, han2019clothflow} pose the problem as that of synthesizing a realistic image of a model from a reference image and an isolated garment image. VITON \cite{viton} uses a Thin-Plate Spline (TPS) based warping method to deform the garment images and maps the warped garment onto the model image using an encoder-decoder refinement module. CP-VTON \cite{cpvton} improves over \cite{viton} using a neural network to regress the transformation parameters of TPS. SieveNet \cite{sievenet} improves over \cite{cpvton, viton} by estimating TPS parameters over multiple interconnected stages and also proposes a conditional layout constraint to better handle pose variation, bleeding and occlusion during texture fusion. ACGPN \cite{acgpn} utilizes a similar layout constraint and also imposes a second-order constraint on TPS warping to preserve local patterns. However, these methods can only model limited geometric changes and often unnaturally deform clothing due to limited degrees of freedom in TPS transformation. ClothFlow~\cite{han2019clothflow} uses a per-pixel appearance flow~\cite{appearanceFlow} (instead of TPS) predicted over multiple cascaded stages, and also utilizes the conditional layout constraint as in \cite{sievenet, acgpn}. Appearance flow ~\cite{appearanceFlow} is used to spatially deform a source scene to a target scene by computing a pixel-wise 2D deformation field. This is conceptually distinct from optical flow and we refer the reader to ~\cite{appflowdiff} for a discussion regarding the difference. The high degree of freedom in per-pixel flow estimation as well as the limited (3D) structural information often results in geometric misalignment and unnatural and bleeding textures. We propose \OURNAME, an end-to-end framework which seeks to preserve the geometric and textural integrity by a combination of gated aggregation of hierarchical flow estimates across pixel-block levels (\GAF) and dense structural priors (\DGP) at various stages of the network.

\begin{figure*}[t!]
\begin{center}
  \includegraphics[height=0.475\textheight]{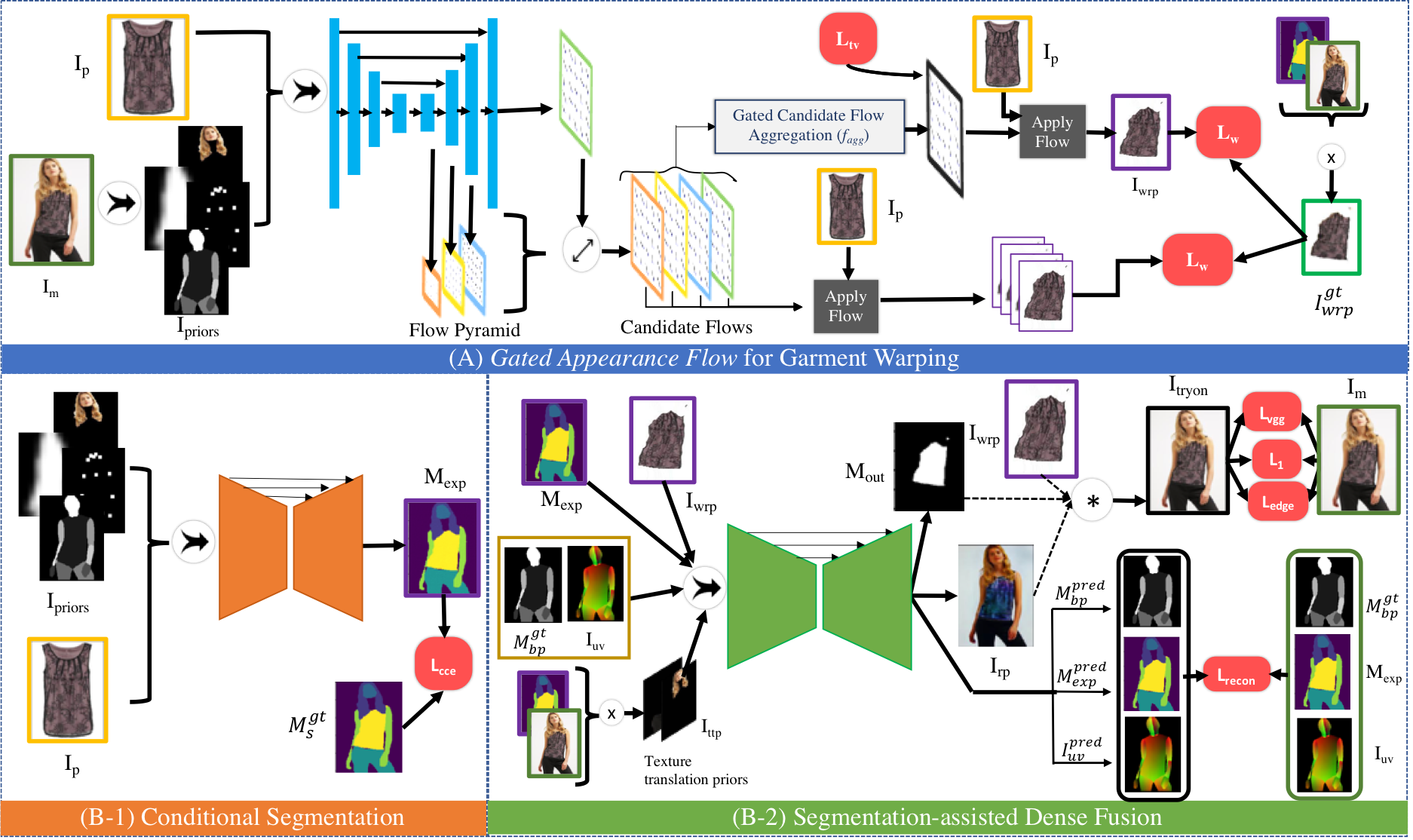}
\end{center}
  \caption{\OURNAME comprises of two modules: A) \underline{Garment Warping} to deform the garment $I_{p}$ to align with model $I_{m}$ and generates warped garment ($I_{wrp}$), and B) \underline{Texture Fusion} which has two sub-steps - i. Conditional Segmentation to predict a post try-on clothing segmentation of the model $M_{exp}$ ii. Dense Fusion which combines the warped garment ($I_{wrp}$) and segmentation mask ($M_{exp}$) to generate the final output ($I_{tryon}$). \textit{\GAF} for garment warping improves textural integrity of $I_{tryon}$ by regularizing the per-pixel flow estimation. Dense geometric priors $I_{priors}$ improves geometric integrity of the try-on output.}
\label{fig:overview}
\end{figure*}

\textbf{3D Pose Representation} The optimal choice of 3D representations for neural networks is an open problem. Recent works in single image 3D reconstruction have explored voxel, point cloud, octree, surface and volumetric representations \cite{varol2018bodynet, loper2015smpl, yao2019densebody, alp2018densepose, zheng2019deephuman, kanazawa2018end, jackson20183d}. Surface based representation methods~\cite{alp2018densepose, yao2019densebody} use UV maps \cite{feng2018joint} to establish dense correspondence between pixels and human body surface. To preserve the geometric integrity (depth-ordering, pose, skin and neckline reconstruction) of the try-on output in our image-based setup, we use dense geometric priors in form of UV maps and body-part segmentation masks obtained from a pre-trained DensePose \cite{densepose}. These priors helps in handling complex poses even under heavy occlusion.

\textbf{Human Pose Transfer} Given a reference image of a person and a target pose, the task is to synthesize an image of the model in the desired pose. \cite{ma2017pose} uses a two stage, guided image-to-image translation network to generate the target. Recent work \cite{siarohin2018deformable, dong2018soft, balakrishnan2018synthesizing, han2019clothflow, li2019dense, grigorev2018coordinate} incorporates spatial deformation from the source to the target for better perceptual quality. ClothFlow \cite{han2019clothflow} predicts a dense appearance flow over multiple interconnected stages using a stacked network to warp source clothing pixels. Dense Intrinsic Flow (DIF) \cite{li2019dense}, introduced a flow regression module to map input and target skeleton poses with 3D appearance flow which it then uses to performs feature warping on the input image and generate a photo-realistic target image. We validate the efficacy of \textit{\GAF} by adapting it for regression of 3D flows in ~\cite{li2019dense}. We note subsequent work in pose transfer ~\cite{hpt-sota} but highlight that ~\cite{li2019dense} is ideal for our objective of validating the efficacy of \textit{\GAFshort}.
\section{Methodology}
\label{sec:method}
\OURNAME takes as input images of a target model ($I_{m}$) and an isolated garment product ($I_{p}$) to generates the \textit{try-on} output $I_{tryon}$, where the target model is wearing the garment. This transformation is composed of two key phases: (\textit{A}) \textbf{Garment Warping} which deforms $I_{p}$ to align with pose of the model in $I_{m}$ and generates $I_{wrp}$, (\textit{B}) \textbf{Texture Fusion} which composes the warped garment $I_{wrp}$ with $I_{m}$ to generate $I_{tryon}$ over two steps: (\textit{B-1}) conditional segmentation and  (\textit{B-2}) segmentation-assisted fusion (as in Figure~\ref{fig:overview}).
\subsection{Garment Warping}
$I_{p}$ is warped based on pose and shape of the target model $I_{m}$ to produce a warped garment image $I_{wrp}$. For this, we propose \textit{\GAF} which estimates per-pixel warp parameters by aggregating candidate estimates predicted across multiple scales (pixel-block sizes).
\vspace{-4mm}
\subsubsection{Enriched Input} Because training triplets where the same model wears two different garments are unavailable, contemporary methods use as input a clothing-agnostic prior of the target model ($I_{m}$) along with the garment $I_{p}$. We extend the conventional binary (1-channel) body shape, (18-channel) pose map and (3-channel) head region used previously ~\cite{acgpn, sievenet, han2019clothflow} with an additional dense (11-channel) body-part segmentation ($M_{bp}^{gt}$) of $I_{m}$ to provide richer structural priors ($I_{priors}$). This subtle enhancement, as we delineate in section~\ref{sec:ablations}, cascades through the network and results in significantly fewer artefacts in the output.

\subsubsection{Gated Appearance Flow} This module predicts per-pixel appearance flow (pixel displacements) for warping the garment image by aggregating candidate flow estimates across multiple scales. The process comprises of first predicting the flow estimates and then aggregating them using a gating mechanism, along with losses that ensure smoothness (and regularity) of the flow predictions.  
\vspace{-2mm}
\paragraph{Multi-scale Appearance Flow Prediction} The backbone network is a 12-layer Skip-Unet \cite{UNet}. 
Given an input RGB image of size ($H, W$), the last $K$ decoding layers are used to predict the candidate flow maps ($f_{l}$ for $l \in \{0, ..., K\}$) such that a predicted map $f_{l}$ is double the size of map $f_{l-1}$. All maps are then interpolated to have identical height and width ($H,W$) generating a pyramid of $K$ \textit{candidate} flow maps that correspond to a structural hierarchy.
\vspace{-2mm}
\paragraph{Appearance Flow Aggregation}
The candidate flows are combined to obtain an aggregate per-pixel appearance flow ($f_{agg}$), using a convolution gated recurrent-network (ConvGRU) ~\cite{conv-GRU} (summarized in figure~\ref{fig:overview}(A)). Intuitively, this is a per-pixel selection process that determine the aggregate flow by gating (allowing or dismissing) pixel flow estimates corresponding to different radial neighborhoods (for the multiple scales). This prevents over-warping of the garment image by regularizing the high degrees of freedom in dense per-pixel appearance flow. We corroborate this position with extensive ablation studies in section ~\ref{sec:ablations-gatedflow} where we propose and contrast several alternative flow aggregation mechanisms. 

\vspace{-4mm}
\paragraph{Garment Image Warping} Next, the aggregate appearance flow map $f_{agg}$ is used to warp the garment image $I_{p}$ and mask $M_{p}$ to obtain the warped image $I_{wrp}$ and the warped binary garment mask $M_{wrp}$ respectively. Additionally, the intermediate flow maps $f_l$ for $ l \epsilon \{0, .., K\}$ are also used to produce intermediate warped images and masks ($I^{l}_{wrp}, M^{l}_{wrp}$).

\vspace{-4mm}
\paragraph{Losses} 
Each of the warped images (final and intermediate) are subject to L1-loss $L_{1}$ and perceptual similarity loss $L_{vgg}$ ~\cite{VGGCNN} with respect to garment regions of the model image. Each predicted warped mask is subject to a reconstruction loss with respect to $M^{gt}_{m}$. The predicted flow-maps are subjected to a total variation loss ($\beta_4 L_{tv}(f_{l})$) to ensure spatial smoothness of flow-predictions. The combined warping loss is defined as $L_{wrp}$ :

\begin{small}
\begin{align}
\begin{split}
L_{wrp} &= L_w(I_{wrp}, M_{wrp}, f_{agg}) \\ &\enskip +  \sum_{l=0}^{l=K}L_w(I^{l}_{wrp}, M^{l}_{wrp}, f_l)
\end{split}
\end{align}
\end{small}
for,
\begin{small}
\begin{align}
\begin{split}
L_w(I,M,f) &=\beta_{1} \|I\odot M, I_m\odot M^{gt}_{m}\|_1 \\
&\enskip + \beta_{2} L_{vgg}(I\odot M, I_m\odot M^{gt}_{m})  \\
&\enskip +\beta_{3} \|M, M^{gt}_{m}\|_1 + \beta_{4}L_{tv}(f)
\end{split}
\end{align}
\end{small}

\vspace{-4mm}
\paragraph{Validation with Human Pose Transfer}
For an extended validation of \GAFshort's efficacy for estimating appearance flows, we use it to regress 3D flows for human pose transfer. The task involves producing an image of a person in a target pose from a reference image. We note that in contrast to virtual try-on where \textit{GAF} is used for warping the \textit{garment} based on the model pose, here it warps the target model pose itself. DIF ~\cite{li2019dense} is a recent method for pose transfer that first regresses on a 3D appearance flow to map input to target pose and then performs feature warping on the input using the flow estimates. We swap-in our proposed \GAFshort for 3D flow regression while retaining the feature warping module of DIF. We observe significant qualitative improvement in the generated image and discuss the results in section~\ref{sec:ablations}. 

\subsection{Texture Fusion}
Once the warped garment ($I_{wrp}$) is obtained, the final try-on output is then generated over two steps (figure~\ref{fig:overview} B-1 and B-2): First, a conditional mask $M_{exp}$ is predicted that corresponds to the clothing segmentation of the target model \textit{after} garment change in try-on. Then, $M_{exp}$ is combined with the warped garment ($I_{wrp}$) and the texture and geometry priors to produce the try-on output ($I_{tryon}$) . 
\vspace{-4mm}
\subsubsection{Conditional Segmentation} The inputs to this module are the garment image ($I_{p}$) and the \DGAR ($I_{priors}$). The $I_{priors}$ encodes the geometry of the target person and is agnostic to the specific garment the model is wearing. This is important to prevent over-fitting as the pipeline is trained on paired data where the input and output are the same images (and hence have the same segmentation mask). The network architecture is a Skip-UNet \cite{UNet} with six encoder and decoder layers and the output, $M_{exp}$, is the 7-channel clothing segmentation mask. 
\vspace{-2mm}
\paragraph{Losses} The module is trained with a weighted cross-entropy loss with respect to the ground-truth garment segmentation mask ($M^{gt}_s$) obtained with a pre-trained human parser (as used in ~\cite{sievenet, acgpn, han2019clothflow}). The weight for skin and background classes are increased (3.0 in our experiments) for better handling of bleeding, and self-occlusion where the pose of the person results in certain parts of the garment or body to remain hidden from view. The loss is expressed as:
\begin{align}
\begin{split}
L_{cs} &= -\frac{1}{n}\sum_n\sum^6_{i=0} w_i P^{gt}_i log( P^{pred}_i)\\
\text{where } w_i &= [3, 1, 1, 1, 3, 1, 1] \text{ for } i \epsilon [0, 6] 
\end{split}
\end{align}

We observe that using the \DGAR improves depth perception and handling of occlusion in $M_{exp}$ which results in try-on outputs with fewer artefacts. We discuss this further in section~\ref{sec:ablations_input_priors}.
\subsubsection{Segmentation-Assisted Dense Fusion}
This stage generates the final try-on output. The network architecture for this stage is also a Skip-UNet \cite{UNet} with six encoder and decoder layers. The network inputs include outputs of the previous stages ($I_{wrp}$ and $M_{exp}$) and texture translation prior ($I_{ttp} = I_{m}*M_{exp}$) representing the non-garment pixels of $I_{m}$. To include the 3D geometry of the model, we also input a dense prior (called \textit{\IUV}) composed of UV map ($I_{uv}$) and body-part segmentation ($M^{gt}_{bp}$) of the target model. We note that $M^{gt}_{bp}$ (\textit{body-part} segmentation) is a function of the body geometry (agnostic of the specific garments) and differs from $M_{exp}$ (or $M^{gt}_s$) (\textit{clothing} segmentation) which is altered with changing garments (both are useful for try-on).
The try-on output ($I_{tryon}$) is defined as:
\begin{dmath}
    I_{tryon} = M_{out} * I_{wrp} + (1 - M_{out}) * I_{rp}
\end{dmath}
where $M_{out}$ and $I_{rp}$ are generated by the network. $M_{out}$ is a composite mask for the garment pixels in try-on output and $I_{rp}$ is a \textit{rendered person} comprising all target model pixels \textit{except} the garment in the try-on output. To preserve structural and geometric integrity of the try-on output, we also constrain the network to reconstruct the input clothing segmentation (as $M^{pred}_{exp}$) and IUV (as $M^{pred}_{bp}, I^{pred}_{uv}$) priors which are unchanged during this step.
\vspace{-4mm}
\paragraph{Losses} $I_{tryon}$ is subject to $L_{1}$, perceptual similarity~\cite{VGGCNN} ($L_{vgg}$) and edge ($L_{edge}$) losses with respect to the model image $I_{m}$. $L_{edge}$ is based on sobel filters ($\nabla_x$ and $\nabla_y$) and improves quality of the reproduced textures. Finally, $M^{pred}_{exp}$, $M^{pred}_{bp}$ and $I^{pred}_{uv}$ are subjected to reconstruction losses against their corresponding network inputs ($M_{exp}$, $M^{gt}_{bp}$ and $I_{uv}$ respectively). This reconstruction loss ($L_{recon}$) combines cross entropy ($L_{cce}$) for the categorical masks ($M^{pred}_{exp}$, $M^{pred}_{bp}$) and smooth $L_{1}$ for the $I^{pred}_{uv}$ map. 
\begin{small}
\begin{align}
\begin{split}
L_{fus} &= \lambda_{1}*\|I_{tryon}-I_{m}\|_1 + \lambda_{2}*L_{vgg}(I_{tryon}, I_{m})\\
&\enskip +  \lambda_{3}*L_{edge}(I_{tryon}, I_{m}) + \lambda_{4}*L_{recon}
\end{split} \raisetag{12pt} 
\end{align}
where,
\small
\begin{align}
\begin{split}
    L_{recon} &= L_{cce}(M^{pred}_{exp}, M_{exp}) + L_{cce}(M^{pred}_{bp}, M^{gt}_{bp})\\  
    &\enskip + \|I^{pred}_{uv} - I_{uv}\|_{smoothL1}
\end{split}
\end{align}
\end{small}

We observe that conditioning texture fusion with these geometric priors via $L_{recon}$ improves quality of try-on output via improved depth perception and structural coherence and explain this effect with evidence in section~\ref{sec:ablations}.

\subsection{Training}
Following a brief warm-up period of $\tau$ steps for the warping and texture fusion modules where they are trained separately, we optimize \OURNAME end-to-end with the following loss function:
\begin{equation}
    L_{total} = \alpha_{1}*L_{wrp} + \alpha_{2}*L_{cs} + \alpha_{3}*L_{fus}
\end{equation}
where $\alpha_1, \alpha_2, \alpha_3$ are scalar hyperparameters.
\section{Experiments}
In this section, we formalise the setup for our experiments with virtual try-on and human pose transfer. 
\vspace{-4mm}
\paragraph{Datasets} For image-based virtual try-on, we use the VITON dataset \cite{viton} to ensure consistence with baseline methods. It contains 19000 images of front-facing female models and corresponding upper-clothing isolated garment images of size 256x192. There are 16253 cleaned pairs, which are split into train and test sets of 14221 and 2032 pairs. We also separate out 500 pairs from the train set into a validation set used exclusively for quantitative analysis. The images in the test set are rearranged into unpaired sets for qualitative evaluation. For human pose transfer, we use in-shop clothes benchmark from the Deep Fashion dataset \cite{deepfashion} which contains 52712 in-shop clothes images and 200000 cross-pose pairs of size 256x256. Following the setup in DIF \cite{li2019dense}, we select 89262 pairs and 12000 pairs for train and test respectively.
\vspace{-4mm}
\paragraph{Implementation Details} All experiments are conducted using Pytorch on Tesla V100 GPUs. For virtual try-on, all modules are trained for 30 epochs with a batch size of 4 and learning rate of 1e-4 using Adam \cite{kingma2014adam}. We set $K=3$ for \textit{\GAF} and the warm-up period $\tau$ is 5 epochs. For human pose transfer, we train the flow regression module for 40 epochs with learning rate=1e-4 using Adam \cite{kingma2014adam} and retain the configuration in \cite{li2019dense} for the feature warping module. Additional hyperparameter details are in the appendix.
\vspace{-4mm}
\paragraph{Evaluation Metrics} For virtual try-on, we use SSIM \cite{seshadrinathan2008unifying}, FID \cite{heusel2017gans} and PSNR \cite{hore2010image} of the warp garment and try-on output. We avoid inception score (IS) following the considerations presented in \cite{barratt2018note}. For human pose transfer, we evaluate performance using SSIM \cite{seshadrinathan2008unifying} and PSNR \cite{hore2010image} to ensure consistency with baselines. We note these metrics are chosen to ensure consistent comparison with prior work.

\vspace{-4mm}
\paragraph{Baselines} For virtual try-on, we compare performance with several recent state-of-the-art methods including CP-VTON \cite{cpvton}, SieveNet \cite{sievenet}, ClothFlow \cite{han2019clothflow}, VTNFP \cite{vtnfp} and ACGPN \cite{acgpn}. For \cite{cpvton, sievenet, acgpn}, we use author provided implementations and perform extensive qualitative and quantitative comparisons.  

\section{Results}
~\label{sec:results}
\begin{figure}
\begin{center}
  \includegraphics[width=\linewidth]{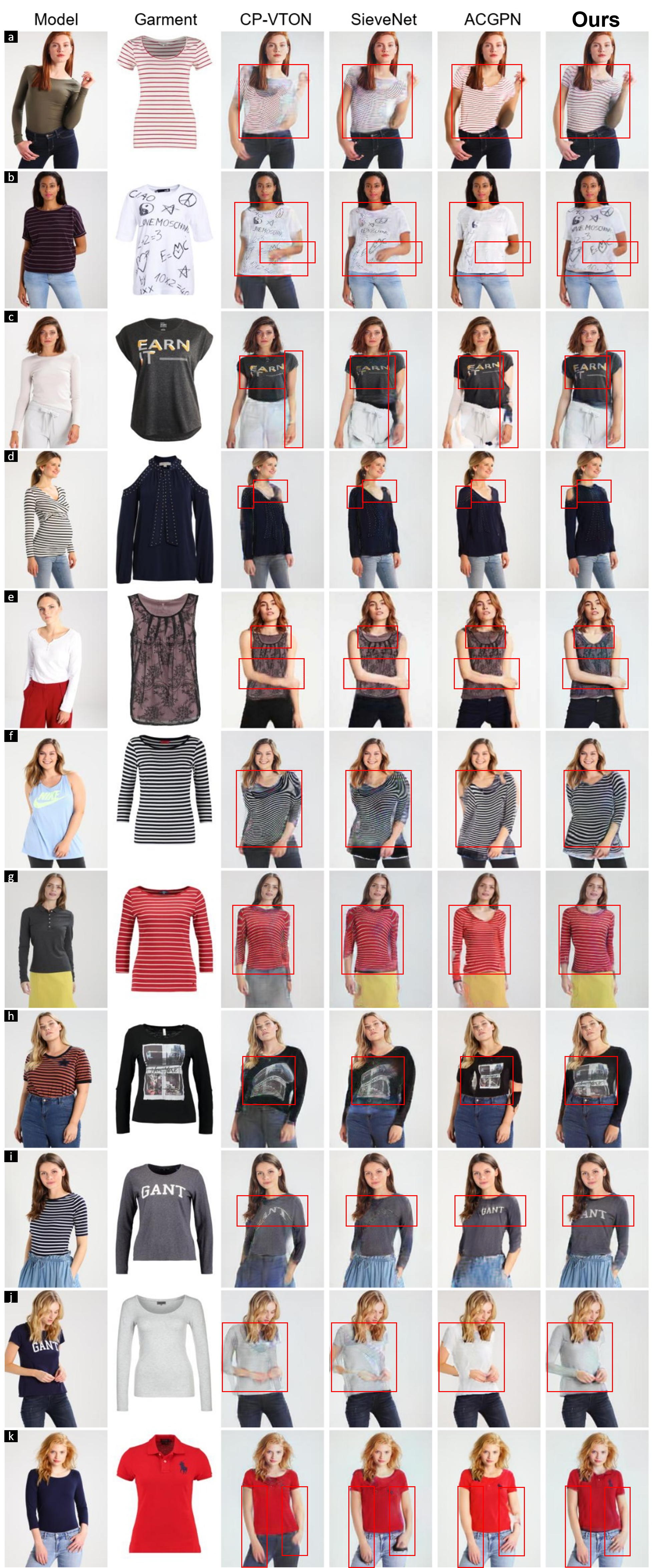}
\end{center}
\caption{Qualitative Comparison of \OURNAME with \cite{cpvton, sievenet, acgpn}. Rows \textbf{1-5} reflect improvements in preserving geometric integrity, and Rows \textbf{6-10}, texture integrity. Please note: (a) Complex poses (b) Depth ordering of body parts (c) Skin generation (d,e) Neckline and shoulder correction (f,g) Pattern (h) Texture, (i) Text, (j) Reduced bleeding across part boundaries. Row \textbf{11} (k) Realistic outline shadows for crisper image quality. (Best viewed with zoom). Please see appendix (\underline{pages 3-5}) for more results.} 
\label{fig:qual1}
\end{figure}

We present quantitative (in Table~\ref{tab:quant-sota}) and qualitative results (Figure~\ref{fig:qual1}) along with a user study which highlight the superiority of \OURNAME over strong baselines.

\begin{table}[h!]
\begin{center}
\begin{tabular}{|c|c|c|c|}
\hline
Method    & SSIM $\uparrow$  & PSNR $\uparrow$  & FID $\downarrow$  \\
\hline
VTNFP \cite{vtnfp}$^{\dagger}$  & 0.803 & -     & -     \\
ACGPN \cite{acgpn}$^{\dagger}$  & 0.845 & -     & -     \\
\hline
CP-VTON \cite{cpvton}   & 0.784 & 21.01 & 30.50 \\
SieveNet \cite{sievenet}  & 0.837 & 23.52 & 26.67 \\
ClothFlow \cite{han2019clothflow} & 0.843 & 23.60 & 23.68  \\
\hline \hline
\textbf{\OURNAME}   & \textbf{0.885} & \textbf{25.46} & \textbf{15.17} \\
\hline
\end{tabular}
\end{center}
\caption{\OURNAME achieves significant improvement over existing baselines. $^{\dagger}$ results may be inferred as indicative as they are transferred from corresponding papers.}
\label{tab:quant-sota}
\end{table}

\vspace{-4mm}
\paragraph{Quantitative Results}
Table~\ref{tab:quant-sota} compares performance of \OURNAME against state-of-the-art baselines for virtual try-on. We report performance for TPS-based baselines ~\cite{cpvton, sievenet} using author provided implementations. In comparison to ~\cite{cpvton, sievenet}, \OURNAME achieves significantly better SSIM of 0.885, PSNR of 25.46 and FID of 15.17, compared to the next best values (SSIM=0.845, PSNR=23.60 and FID=23.68). We note that \OURNAME with \textit{\GAFshort} significantly outperforms \textit{ClothFlow} ~\cite{han2019clothflow} which uses vanilla per-pixel appearance flow based warping for the garment image. Note that the official code for ClothFlow \cite{han2019clothflow} was not available, we implement it as described and reproduce stated SSIM values.

\vspace{-4mm}
\paragraph{Qualitative Results}
Figure~\ref{fig:qual1} illustrates qualitative comparison with SieveNet~\cite{sievenet}, CP-VTON~\cite{cpvton} and ACGPN~\cite{acgpn}, the baselines with available code implementations. We contrast the try-on outputs along varying dimensions of quality. These include factors that determine the realism of the generated image as a whole as well as the local geometry, colors and patterns. 

Rows (1-5) demonstrate improvement in \textit{geometric integrity} - the accurate representation of the geometry of the target model, the garment, and their interaction in the try-on output. Specifically, we observe that \OURNAME improves the handling of \textit{extreme pose} (row 1), \textit{depth-ordering} of body parts, especially hands and neck region (row 2), \textit{skin generation} for correct visibility of target garment and human skin (row 3) and \textit{neckline reproduction \& shoulder correction} in coherence with garments structure (row 4, 5). We highlight the improved neckline reproduction and depth-ordering in row 5 where none of the baselines are able to disambiguate front and back of the garment neckline.

Rows (6-10) demonstrate improvement in \textit{texture integrity} which is concerned with accurate reproduction of patterns and colors of inshop garments in try-on output, and the handling of related artefacts. Specifically, we observe that \OURNAME improves the reproduction of \textit{pattern and texture} (stripes in row 6, 7), \textit{print design} of the garment (graphic in row 8), \textit{text} written on garment (row 9) and prevents color bleeding across part boundaries (row 10). 

Shadows and highlights in the generated image, especially along the boundaries of body parts, are also important to correctly represent the dynamics of the actual scene. Row 11 demonstrates improvement along this dimension. 



\vspace{-4mm}
\paragraph{User Study}
We conduct a survey with 70 volunteers from 3 continents, 5 countries, 10 institutions across diverse age, gender and occupations. As in ~\cite{acgpn}, we use \textit{pairwise comparison} where each user is shown 100 distinct result \textit{pairs} randomly sampled from 2000 test set results. Each pair consists of one \OURNAME result and the other sampled from the results of (one of three) baselines (~\cite{acgpn, sievenet, cpvton}). The in-shop garment and target model images are also shown for each result pair. Every volunteer is asked to select the best output of the two in each result pair in unlimited time. Results in Table~\ref{tab:human-study} show overwhelmingly clear preference for \OURNAME in \textbf{\textit{all}} pairwise comparisons. 

\begin{table}
\begin{center}
\begin{tabular}{|c|c|c|}
\hline
Baseline & Prefer Baseline  &  Prefer \OURNAME \\
\hline
\hline
CP-VTON \cite{cpvton}   & 8\% & \textbf{92\%} \\
\hline
SieveNet \cite{sievenet}  & 15\%  & \textbf{85\%} \\
\hline
ACGPN  \cite{acgpn}   & 29\%  & \textbf{71\%} \\
\hline
\end{tabular}
\end{center}
\caption{Survey results for gauging the human preference of \OURNAME over competing baselines. The percentage indicates the ratio of images which are voted to be better than the compared method.}
\label{tab:human-study}
\end{table}

\begin{table*}[h!]
\begin{center}
\begin{tabular}{|c|c|c|c|c|c|c|}
\hline
\multicolumn{2}{|c|}{Configuration}            & \multicolumn{2}{c|}{Warp Garment ($I_{wrp}$)} & \multicolumn{3}{c|}{Try-On Output ($I_{tryon}$)} \\
\hline
\textit{Garment Warping} & \textit{Texture Fusion}         & \textit{SSIM $\uparrow$}           & \textit{PSNR $\uparrow$}          & \textit{SSIM $\uparrow$}      & \textit{PSNR $\uparrow$}      & \textit{FID $\downarrow$}      \\
\hline \hline
ClothFlow ~\cite{han2019clothflow}  & BaseFuse                       & 0.835          & 20.54         & 0.843     & 23.60     & 23.68     \\ \hline 
\GAFshort     & BaseFuse                        & \textbf{0.871} & \textbf{23.14} & \textbf{0.865}  & \textbf{24.47} & \textbf{18.89}     \\ \hline\hline
\multicolumn{7}{|c|}{Various Gating Approaches for Flow Aggregation}\\
\hline
Residual Gating   & BaseFuse                        & 0.856          & 22.09         & 0.855     & 24.11     & 21.64     \\ \hline
LSTM    & BaseFuse                        & 0.862          & 22.56         & 0.860     & 24.33     & 18.89     \\ \hline
ConvGRU  (\GAFshort)    & BaseFuse                        & 0.871          & 23.14         & 0.865     & 24.47     & 18.89     \\ \hline
\hline
\multicolumn{7}{|c|}{Loss Functions}\\
\hline
\GAFshort (w/ $I_{priors}$)     & BaseFuse + $L_{edge}$             & 0.871          & 23.28         & 0.875     & 25.02     & 19.39     \\ \hline
\GAFshort (w/ $I_{priors}$)     & BaseFuse + $L_{edge}$ + $L_{recon}$ & 0.871          & 23.28         & 0.876     & 25.12     & 18.74     \\ \hline
\hline
\multicolumn{2}{|c|}{\textbf{\OURNAME (end-to-end training)}}     & \textbf{0.871}          & \textbf{23.28}         & \textbf{0.885}     & \textbf{25.46}     & \textbf{15.17}   \\
\hline
\end{tabular}
\end{center}
\caption{Ablation studies for various design choices for garment warping and texture fusion in \OURNAME. \textit{BaseFuse} is the texture fusion network trained without $L_{edge}$ and $L_{recon}$. }. 
\label{tab:ablations-tryon}
\end{table*}
\section{Ablation Studies}
\label{sec:ablations}
In this section, we analyse the impact of different contributions of \OURNAME and summarize results in Table~\ref{tab:ablations-tryon}.

\subsection{Gated Appearance Flow (GAF)}
\label{sec:ablations-gatedflow}
First, we demonstrate the impact of \GAFshort for garment image warping by comparing it to an existing per-pixel appearance flow based warping technique proposed in ClothFlow \cite{han2019clothflow}. Next, to justify our choice of using a ConvGRU layer for aggregating hierarchical candidate appearance flow estimates, we propose alternate flow-aggregation schemes and report comparison with ConvGRU.

\vspace{-4mm}
\paragraph{ClothFlow and \GAFshort}
Rows 1 and 2 of Table \ref{tab:ablations-tryon} compare the use of per-pixel appearance flow for garment image warping as described in \cite{han2019clothflow} with the proposed gated aggregation of hierarchical flow estimates (\GAFshort). \GAFshort clearly outperforms the vanilla warping method corroborating our position that gated aggregation yields superior results both for the warping stage as well as for the try-on output. 
\vspace{-4mm}
\paragraph{Design choices for \GAFshort}
In rows 3, 4 and 5 of Table \ref{tab:ablations-tryon}, we compare following schemes for gated aggregation - \underline{\textit{i) Using Residual Gating}} to perform residual sum (operation from ~\cite{tirg}) on flow estimates of the last two decoding layers. \underline{\textit{ii) Using ConvLSTM}} for the flow estimate aggregation over three layers (3 scales), and \underline{\textit{iii) Using ConvGRU}} for aggregating flow estimates. The results indicate clearly that using \textit{ConvGRU} for gated aggregation produces the \textit{best results of the three} and \textit{hence is used in GAF}.

Further, we note that all three aggregation schemes significantly outperform \textit{ClothFlow} on metrics for both the warped garment and try-on output. For instance, \textit{ConvGRU} improves the warp garment SSIM (from 0.835 to 0.871) and PSNR (from 20.54 to 23.14) against ClothFlow ~\cite{han2019clothflow}. We note that this benefit translates to the try-on output where we observe consistent gains in SSIM (from 0.843 to 0.865), PSNR (from 23.60 to 24.47) and FID (from 23.48 to 18.89).

\begin{figure}
\begin{center}
  \includegraphics[width=0.9\linewidth]{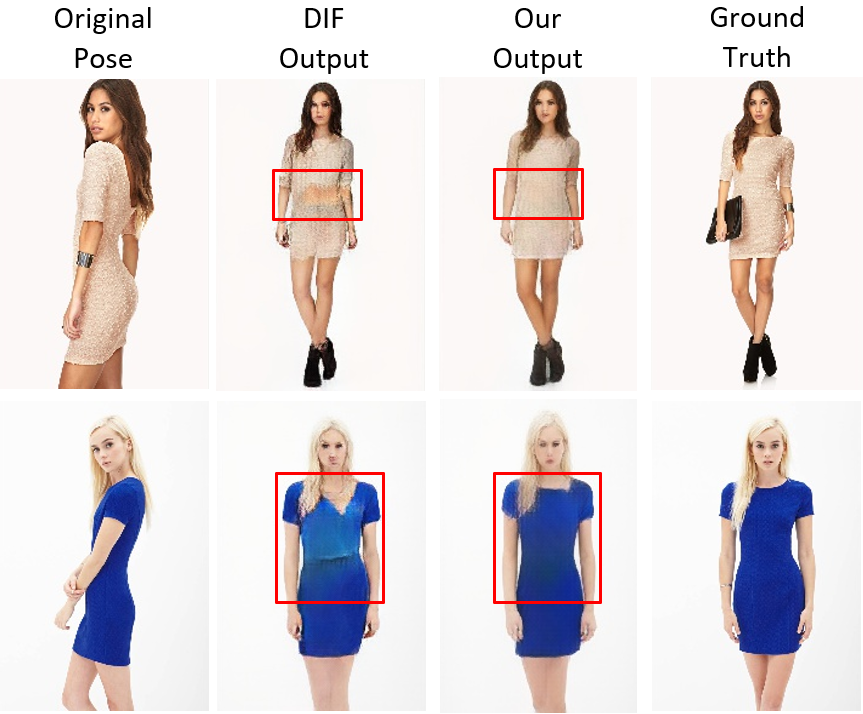}
  
\end{center}
  \caption{Using \textit{\GAFshort} for flow regression in pose transfer improves skin generation (row 1) and reduces bleeding (row 2).}
\label{fig:pose-transfer}
\end{figure}

\vspace{-4mm}
\paragraph{\GAFshort in Human Pose Transfer}
As an additional test of the efficacy of the proposed appearance flow-aggregation, we adapt it for flow-regression for task of Human Pose Transfer, building upon baseline DIF \cite{li2019dense}. This results in both qualitative (Figure \ref{fig:pose-transfer}) and quantitative (Table \ref{tab:hpt-flowregress}) improvements in the pose-transfer output. Figure~\ref{fig:pose-transfer} present evidence to show significantly improved skin generation (row 1), texture (row 2) and reduced bleeding (row 1, 2) in the generated image. We corroborate this with results in Table~\ref{tab:hpt-flowregress} which indicates considerable improvement in SSIM (from 0.778 and 0.791) and PNSR (from 18.59 to 19.26). We also note the significant gain over ClothFlow~\cite{han2019clothflow}, which also uses flow regression, as a validation of the efficacy of GAF.

\begin{table}[h!]
    \begin{center}
    \begin{tabular}{|c|c|c|}
    \hline
    Method      & SSIM $\uparrow$  & PSNR $\uparrow$ \\
    \hline \hline
    DSC  \cite{siarohin2018deformable}       & 0.756 & -     \\
    PG2 \cite{ma2017pose}        & 0.762 & -     \\
    ClothFlow \cite{han2019clothflow}    & 0.771 & -     \\
    VUnet \cite{esser2018variational}      & 0.786 & -     \\
    DIF \cite{li2019dense}        & 0.778 & 18.59 \\
    \hline \hline
    \textbf{Ours} & \textbf{0.791} & \textbf{19.26} \\
    \hline
    \end{tabular}
    \end{center}
    \caption{Using \GAFshort for flow regression improves the quality of generated image in human pose transfer}
    \label{tab:hpt-flowregress}
\end{table}

\begin{figure}
\begin{center}
  \includegraphics[width=\linewidth]{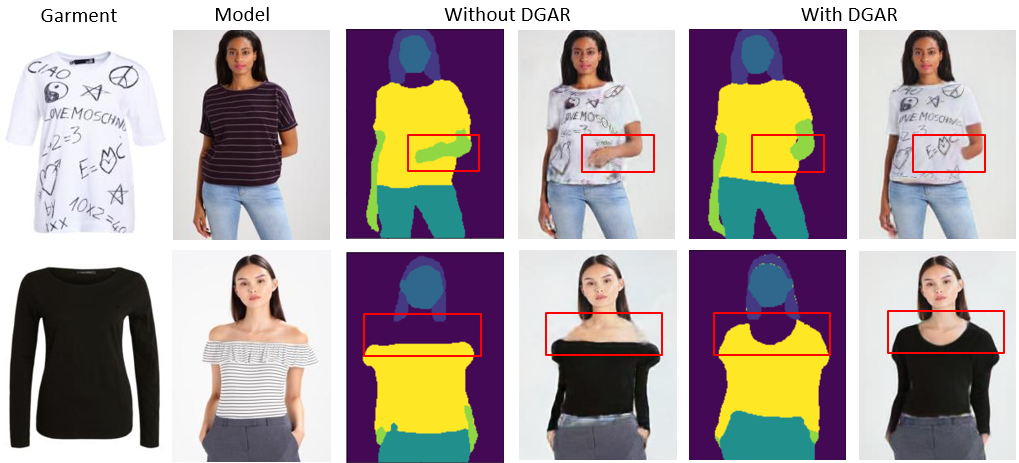}
\end{center}
  \caption{Using the dense garment-agnostic representation (\DGARshort) for conditional segmentation improves depth-perception (row 1), skin and neckline generation (row 2).}
\label{fig:conditional-seg}
\end{figure}

\subsection{Input Priors, Losses and Training}
\label{sec:ablations_input_priors}
\paragraph{\DGAR} ($I_{priors}$) is proposed as structural priors for garment warping and conditional segmentation. Figure \ref{fig:conditional-seg} shows that this improves depth perception, skin generation (row 1) and neckline reconstruction (row 2) in the try-on output. We note similar improvements during garment warping (qualitative in appendix) which is corroborated through increase in PSNR of the warp garment (row 5 vs 6 in Table~\ref{tab:ablations-tryon}).
\vspace{-2mm}
\paragraph{\IUV} composed of UV projection map ($I_{uv}$) and body-part segmentation ($M^{gt}_{bp}$) are used to encode the 3D geometry of the target model during texture fusion.
The $\OURNAME$ network is trained to reconstruct these priors along with the try-on output ($I_{tryon}$). Figure~\ref{fig:iuv-priors} shows that conditioning on these \IUV via the reconstruction loss ($L_{recon}$) improves generation of neckline, skin (row 1) and depth perception (row 2) in the output. This is corroborated through improved PSNR (25.02 to 25.12) and FID (19.39 to 18.74) of the try-on output (row 6 vs 7 in Table~\ref{tab:ablations-tryon}).

\begin{small}
\begin{figure}
\begin{center}
  \includegraphics[width=0.8\linewidth]{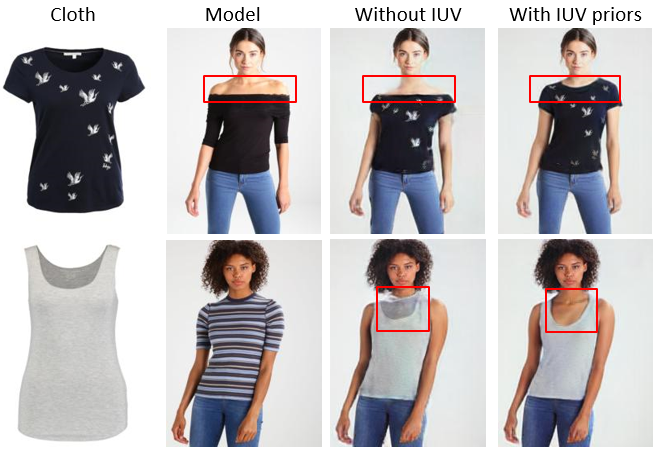}
\end{center}
  \caption{IUV priors during texture fusion improves the neckline (row 1), depth perception and skin generation (row 2)}
\label{fig:iuv-priors}
\end{figure}
\end{small}

\vspace{-2mm}
\paragraph{Edge Loss} ($L_{edge}$) based on Sobel filters is used to better preserve high frequency details during texture fusion. Table~\ref{tab:ablations-tryon} show that this improves SSIM (from 0.865 to 0.875) and PSNR (from 24.47 to 25.02) of try-on output. 
\vspace{-5mm}
\paragraph{End-to-end Fine Tuning} The end-to-end fine-tuning of the entire \OURNAME network (including the warping and texture fusion modules) improves SSIM (0.876 to 0.885), PSNR (25.12 to 25.46) and FID (from 18.74 to 15.17) of the try-on output as indicated in Table~\ref{tab:ablations-tryon} (row 7 vs 8).


\section{Conclusion}
We introduce \OURNAME, an end-to-end try-on framework, which utilizes a combination of gated aggregation of hierarchical flow estimates (\GAF) and dense geometric priors (\DGARshort and \IUV) to reduce undesirable output artefacts. We highlight effectiveness of \OURNAME through comparisons with state-of-the-art and detailed ablation studies. We also validate the efficacy of \GAFshort as a general technique by applying it to human pose transfer. 


{\small
\bibliographystyle{ieee_fullname}
\bibliography{main}
}

\end{document}